\documentclass[11pt]{article}
\usepackage{acl}
\usepackage{times}
\usepackage{latexsym}
\usepackage{microtype}
\usepackage{inconsolata}
\usepackage{graphicx}
\usepackage{comment}
\usepackage{amsmath,amssymb,booktabs}
\usepackage{tikz}
\usetikzlibrary{shapes.geometric, arrows.meta, positioning, fit, backgrounds, calc}
\usepackage{float}
\usepackage{array}
\usepackage{enumitem}
\usepackage{multirow}
\usepackage{hyperref}
\usepackage{fontspec}

\newfontfamily{\hindifont}[
    Path      = ./,
    Extension = .ttf,
    UprightFont = NotoSerifDevanagari-Regular
]{NotoSerifDevanagari}

\newfontfamily{\malayalamfont}[
    Path      = ./,
    Extension = .ttf,
    UprightFont = NotoSerifMalayalam-Regular
]{NotoSerifMalayalam}

\newfontfamily{\cjkfont}[
    Path      = ./,
    Extension = .otf,
    UprightFont = NotoSansCJKsc-Regular
]{NotoSansCJKsc}
\title{Conceptual Networks for Cross‑Linguistic Idiomatic Expressions: A Feature‑Based Graph Approach}
\author{
  \begin{tabular}{c}
    \textbf{Kiran Pala}\thanks{Corresponding author.} \\
    University of Eastern Finland, Finland \\
    \texttt{kiran.pala@uef.fi}
  \end{tabular}
  \\[2em]
  \begin{tabular}{c@{\hskip 3em}c}
    \textbf{Punam Silu} & \textbf{Luxin Yu} \\
    Indian Institute of Technology Ropar, India & University of Eastern Finland, Finland \\
    \texttt{punam.24hsz0007@iitrpr.ac.in} & \texttt{luxinyu@uef.fi}
  \end{tabular}
}

\begin{document}
\raggedbottom
\maketitle

\begin{abstract}
We present an interpretable network-based framework for representing idiomatic and figurative meaning across eight typologically diverse languages, totaling 160 conventional expressions, the large majority of which are idiomatic. Each expression is annotated with binary conceptual features (containment, concealment, emotional, social, etc.) derived from cognitive-linguistic theory, and pairwise Jaccard similarities define a weighted graph. Community detection reveals that idioms cluster by conceptual schema rather than by language, producing a structure consistent with cognitive-linguistic predictions. The conceptual network captures unique semantic information not present in distributional embeddings, can be scaled via automatic annotation with LLMs, improves downstream idiom detection, and remains robust when enriched with corpus frequencies. Cross-lingual transfer experiments show that conceptual proximity alone can identify acceptable translation equivalents across five language families, with substantial gains over embedding-based baselines. Ablation studies demonstrate that all three feature dimensions---schemas, roles, and valence---contribute non-redundantly to both the network's organizational properties and its performance on idiom detection, and that specific graph-derived signals (community membership, neighbor similarity) are particularly informative. The framework offers an interpretable, cross-linguistically stable representation of idiomatic meaning, combining theoretical grounding with practical utility.
\end{abstract}

\section{Introduction}
Idioms such as \textit{spill the beans} or \textit{kick the bucket} pose a fundamental challenge for semantic representation because their figurative meaning cannot be reliably derived from literal constituents. Cognitive linguistics has shown that many idioms are motivated by a small set of image schemas---e.g., \textsc{containment}, \textsc{concealment}---that recur across typologically diverse languages \cite{Kovecses2005}. However, the degree of conceptual overlap between idioms is graded rather than all-or-none, and modeling this graded structure can both advance theoretical understanding and improve natural language processing applications that require handling of non-literal language. Interpretable representations are particularly valuable for NLP tasks where model decisions must be understood and trusted. By grounding representation in cognitive theory, our framework offers both interpretability and cross-lingual stability, complementing black‑box embedding models.

Most computational models of idioms rely on surface lexical forms through distributional semantics or contextual embeddings \cite{Fazly2013}. While powerful, such approaches often miss the deep conceptual structure that underlies idiomaticity across languages, and they provide little insight into the semantic organisation of idioms. Furthermore, existing studies predominantly focus on a small number of well-resourced languages, neglecting the typological diversity that could reveal universal aspects of idiom organisation. Large language models can occasionally identify idioms, but they do not offer a structured conceptual space in which idiomatic meaning can be systematically compared across languages.

We introduce a network-based model of 160 idioms from eight languages representing five language families: Indo-European (English, Hindi, Bagri, Spanish), Uralic (Finnish), Japonic (Japanese), Sino-Tibetan (Chinese), and Dravidian (Malayalam). This selection spans agglutinative, fusional, and isolating morphologies; SVO and SOV word orders; and five distinct writing systems. The inclusion of Bagri and Malayalam adds crucial data points rarely present in idiom studies. The model represents each idiom by a small set of binary conceptual features grounded in cognitive-linguistic theory, and constructs a weighted graph using the Jaccard coefficient as edge weight. We demonstrate that the network organises idioms by conceptual schema rather than by language, and we evaluate its properties through the following complementary analyses: (1) community structure and comparison with multilingual distributional embeddings, (2) downstream idiom detection task evaluation, (3) ablation analyses on feature dimensions, (4) automatic feature extraction via LLMs, (5) dynamic enrichment with corpus frequencies, and (6) cross-linguistic transfer. Together, these analyses show that the conceptual network is a robust, interpretable, and practically useful representation of idiomatic meaning.

\section{Related Work}
\textbf{Cognitive Linguistics and Idioms.}
Conceptual metaphor theory \cite{LakoffJohnson1980} proposed that human thought is structured by a relatively small set of image schemas, which are then realized in language. Gibbs \cite{Gibbs1992} showed that speakers' mental imagery for idioms is highly consistent and reflects underlying conceptual metaphors. Cross-linguistic studies \cite{Piirainen2012} identified widespread idioms that share the same conceptual structure across European and Asian languages, suggesting a universal basis for many idiomatic expressions. These insights motivate our annotation scheme, which captures both universal schemas and language-specific roles.

\textbf{Computational Models of Idioms.}
In NLP, idioms have been modelled using distributional vectors \cite{Fazly2013}, often focusing on detecting whether an expression is used literally or figuratively. However, such models rely on surface co-occurrence statistics and may fail to capture deep conceptual similarities across languages. Graph-based semantic models \cite{SteyversTenenbaum2005} have revealed small-world structure and cognitive correlates, but they have rarely been applied to idioms. Our framework bridges this gap by constructing a conceptual network from theory-grounded features. Existing resources for idiomatic expressions are heavily skewed towards English and a few European languages; our multilingual corpus, low‑resource inclusive dataset addresses this imbalance.

\section{Methodology}
\subsection{Idiom Selection and Corpus Verification}
We compiled 20 conventional figurative expressions (hereafter referred to as idioms) from each of English, Hindi, Bagri, Finnish, Japanese, Spanish, Chinese, and Malayalam, resulting in 160 entries. This broad, cognitively motivated usage of the term \textit{idiom} encompasses not only core phrasal idioms but also proverbs, conventional metaphors, and other fixed expressions whose meaning is not fully compositional; all share the property of being culturally entrenched figurative units, which is the phenomenon of interest here. All expressions were verified by at least two native speakers and checked for frequency in appropriate reference corpora: COCA \cite{Davies2008coca} (English), EMILLE \cite{Baker2002emille} (Hindi), a custom Bagri elicitation set was used, since Bagri (a Rajasthani Indo-Aryan variety) has no standardized written corpus and these figurative expressions were confirmed with three native speakers (all collected with informed consent and anonymized), Suomi24 \cite{Bartis2015suomi24} (Finnish), BCCWJ \cite{Maekawa2014bccwj} (Japanese), CREA \cite{RAE2008crea} (Spanish), BCC/CCL \cite{Xun2016bcc} (Chinese), and the Malayalam Web Corpus \cite{Kilgarriff2010mlwac}. The idioms collected for this study cover common conceptual domains such as communication, emotion, secrecy, and social interaction. They include both cross-linguistically shared metaphorical patterns and idiomatic expressions that are specific to particular languages and cultural contexts. The full dataset will be released upon publication.

\subsection{Feature Annotation Scheme}
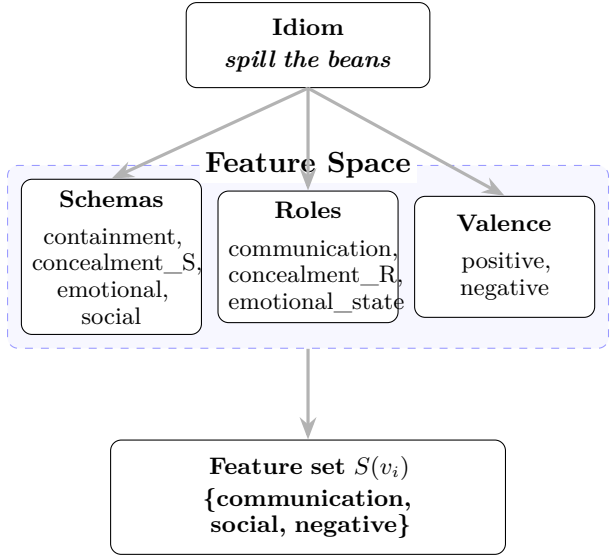
\begin{figure}[H]
\centering
\begin{tikzpicture}[
    idiombox/.style = {draw, rounded corners, fill=white!15, text width=2.8cm, align=center, font=\small\bfseries, inner sep=6pt},
    dimbox/.style   = {draw, rounded corners, fill=white!10, text width=2.1cm, minimum height=1.6cm, align=center, font=\footnotesize, inner sep=4pt},
    outbox/.style   = {draw, rounded corners, fill=white!10, text width=4.8cm, align=center, font=\footnotesize\bfseries, inner sep=6pt},
    arr/.style      = {-{Stealth}, thick, color=gray!60, line width=1.2pt},
]
\node[idiombox] (idiom) at (0,0) {Idiom\\[2pt] \textit{spill the beans}};
\node[dimbox]   (schema)  at (-2.6,-2.8) {\textbf{Schemas}\\[4pt] containment,\\ concealment\_S,\\ emotional, social};
\node[dimbox]   (role)    at ( 0.0,-2.8) {\textbf{Roles}\\[4pt] communication,\\ concealment\_R,\\ emotional\_state};
\node[dimbox]   (valence) at ( 2.6,-2.8) {\textbf{Valence}\\[4pt] positive,\\ negative};

\begin{scope}[on background layer]
  \node[draw=blue!40, dashed, rounded corners, inner sep=5pt, fill=blue!3, fit=(schema)(role)(valence)] (fspace) {};
\end{scope}

\node[font=\bfseries, fill=white, inner sep=2pt] at (fspace.north) {Feature Space};

\draw[arr] (idiom.south) -- (schema.north);
\draw[arr] (idiom.south) -- (role.north);
\draw[arr] (idiom.south) -- (valence.north);
\node[outbox, below=1.2cm of fspace] (out) {Feature set $S(v_i)$\\[2pt] \{communication, social, negative\}};
\draw[arr] (fspace.south) -- (out.north);
\end{tikzpicture}

\caption{Feature annotation schema. An idiom is decomposed into three conceptual dimensions.}
\label{fig:schema}
\end{figure}

Each idiom was manually annotated with binary features drawn from three independent conceptual dimensions, as depicted in Figure~\ref{fig:schema}. The dimensions are grounded in cognitive-linguistic theory and were chosen to capture the core conceptual and functional properties that distinguish idiomatic meanings.

\vspace{1em}

\begin{itemize}[nosep]
\item \textbf{Conceptual schemas}: containment, concealment\_S, emotional, social. These image-schematic structures arise from recurring patterns of bodily experience and have been extensively documented as the conceptual building blocks of metaphorical thought \cite{LakoffJohnson1980}. \textit{Containment} reflects the bounded region schema (a container with an inside and an outside), \textit{concealment\_S} the schema of hiding or covering, \textit{emotional} the mapping of internal affective states onto physical sensations, and \textit{social} the schema of interpersonal interaction and group membership. An idiom receives the containment label if its figurative meaning involves keeping something within a bounded space (e.g., \textit{bottle up your feelings}), concealment if it implies deliberate hiding (e.g., \textit{sweep it under the rug}), emotional if it directly expresses or describes an affective state (e.g., \textit{wear your heart on your sleeve}), and social if it inherently involves an interpersonal dynamic (e.g., \textit{blow the whistle}).
\item \textbf{Functional roles}: communication, concealment\_R, emotional\_state. This dimension captures the pragmatic function that the idiom typically serves in discourse, following the insight that many idioms are not merely descriptive but perform specific communicative acts \cite{Gibbs1992}. \textit{Communication} applies when the idiom's primary function is to convey or elicit information (e.g., \textit{spill the beans}), \textit{concealment\_R} when it describes the act of hiding information or intentions (e.g., \textit{keep it under your hat}), and \textit{emotional\_state} when it ascribes a specific emotion to an experiencer (e.g., \textit{down in the dumps}). An idiom can serve multiple functions simultaneously; for instance, \textit{bite your tongue} both describes concealment of speech and signals an emotional state of restraint.
\item \textbf{Affective valence}: positive, negative. This dimension reflects the evaluative polarity of the idiom's figurative meaning. \textit{Positive} valence is assigned when the idiom denotes a desirable, pleasant, or socially approved outcome (e.g., \textit{get it off your chest}, \textit{bury the hatchet}), while \textit{negative} valence indicates an undesirable, unpleasant, or socially disapproved situation (e.g., \textit{skeleton in the closet}, \textit{sweep it under the rug}). Valence was assessed independently of the literal reading of the words; for example, \textit{swallow your pride} is tagged as negative even though \textit{pride} itself can be positive, because the overall figurative act describes an unpleasant self-suppression.
\end{itemize} 

The dimensions are not mutually exclusive; an idiom can be associated with multiple features. For example, \textit{spill the beans} is tagged \texttt{\{communication, social, negative\}} because it describes revealing a damaging secret in a social context, while \textit{keep it under your hat} receives \texttt{\{containment, concealment\_R, communication, social\}} as it combines the image of a covered container with the communicative act of secret-keeping. The annotation scheme was designed with clear diagnostic criteria—for each feature, a decision tree with prototypical and borderline examples was provided to annotators—and was applied by three trained linguists with expertise in semantics. Inter-annotator agreement was high (Fleiss' $\kappa > 0.78$ for all dimensions), and all disagreements were resolved through consensus discussion. Table~\ref{tab:examples} shows a representative sample across all eight languages.

\begin{table}[H]
\centering
\caption{Example idioms with conceptual features.}
\label{tab:examples}
\small
\begin{tabular}{@{}p{1.5cm}p{1.8cm}p{2.0cm}p{1.6cm}@{}}
\toprule
Lang. & Idiom & Gloss & Feature set \\
\midrule
English & \textit{spill the beans} & --- & \{comm., social, neg.\} \\
Spanish & \textit{guardar un secreto bajo llave} & keep a secret under lock and key & \{cont., conc.\_R, comm.\} \\ \\
Finnish & \textit{haudata asia} & bury the matter & \{conc.\_S, emot.\} \\ \\
Hindi & {\hindifont पेट में बात रखना} & keep matter in stomach & \{cont., conc.\_R, emot.\} \\ \\
Malayalam & {\malayalamfont മനസ്സിൽ സൂക്ഷിക്കുക} & keep in mind & \{cont., conc.\_R, emot.\} \\ \\
Bagri & {\hindifont अकल बिना ऊंट उभाणा फिरै} & Camels roam barefoot without wisdom & \{social, communication, neg.\} \\ \\
Japanese & {\cjkfont 中を見せる} & show one's insides & \{emot., comm., social\} \\ \\
Chinese & {\cjkfont 守口如瓶} & guard mouth like bottle & \{cont., conc.\_R, comm.\} \\
\bottomrule
\end{tabular}
\end{table}

\subsection{Network Construction}
For each pair of idioms \((v_i, v_j)\), the edge weight is the Jaccard coefficient:
\[
w_{ij} = \frac{|S(v_i) \cap S(v_j)|}{|S(v_i) \cup S(v_j)|}.
\]
Jaccard directly quantifies proportional overlap, is bounded, symmetric, and computationally minimal—a resource-rational choice consistent with models of similarity \cite{Tversky1977,LiederGriffiths2020}. The \(160\times160\) matrix \(\mathbf{W}=[w_{ij}]\) defines a weighted graph \(G=(V,E)\); edges with \(w_{ij}>0\) are retained (density 0.37).

\section{Results}
\subsection{Community Structure and Topological Properties}
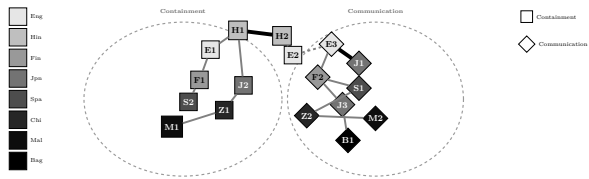
\begin{figure}[H]
\centering
\resizebox{\columnwidth}{!}{%
\begin{tikzpicture}[
    lang/.style={circle, draw=black, minimum size=0.9cm, inner sep=0pt,
                 font=\footnotesize\bfseries},
    % Grayscale gradient: white → black
    en/.style={fill=black!10,  text=black},
    hi/.style={fill=black!25,  text=black},
    fi/.style={fill=black!40,  text=black},
    ja/.style={fill=black!55,  text=white},
    es/.style={fill=black!70,  text=white},
    zh/.style={fill=black!85,  text=white},
    ma/.style={fill=black!95,  text=white},
    bg/.style={fill=black!100, text=white},
    con/.style={regular polygon, regular polygon sides=4, draw=black, inner sep=0pt},
    com/.style={diamond, draw=black, inner sep=0pt},
    edge/.style={draw=black!50, line width=2pt},
    thick edge/.style={draw=black, line width=4pt},
]

% ===== Galaxy ellipses (background) =====
\begin{scope}[on background layer]
  % Containment galaxy
  \draw[black!40, dashed, very thick] (-3,0) ellipse (3.6cm and 2.8cm);
  % Communication galaxy
  \draw[black!40, dashed, very thick] (4,0) ellipse (3.2cm and 2.8cm);
\end{scope}

% Galaxy labels above the clusters
\node[font=\tiny\bfseries,text=black!60] at (-3, 3.2) {Containment};
\node[font=\tiny\bfseries,text=black!60] at (4, 3.2)   {Communication};

% ===== Containment cluster =====
\node[lang, hi, con] (H1)  at (-1.0, 2.5) {H1};
\node[lang, hi, con] (H2)  at ( 0.6, 2.3) {H2};
\node[lang, en, con] (E2)  at ( 1.0, 1.6) {E2};
\node[lang, en, con] (E1)  at (-2.0, 1.8) {E1};
\node[lang, zh, con] (Z1)  at (-1.5,-0.4) {Z1};
\node[lang, ma, con] (M1)  at (-3.4,-1.0) {M1};
\node[lang, ja, con] (J2)  at (-0.8, 0.5) {J2};
\node[lang, fi, con] (F1)  at (-2.4, 0.7) {F1};
\node[lang, es, con] (S2)  at (-2.8,-0.1) {S2};

% ===== Communication cluster =====
\node[lang, fi, com] (F2)  at ( 1.9, 0.8) {F2};
\node[lang, es, com] (S1)  at ( 3.4, 0.4) {S1};
\node[lang, ja, com] (J1)  at ( 3.4, 1.3) {J1};
\node[lang, en, com] (E3)  at ( 2.4, 2.0) {E3};
\node[lang, ma, com] (M2)  at ( 4.0,-0.7) {M2};
\node[lang, zh, com] (Z2)  at ( 1.5,-0.6) {Z2};
\node[lang, ja, com] (J3)  at ( 2.8,-0.2) {J3};
% ----- Bagri: moved down to avoid overlap -----
\node[lang, bg, com] (BG1) at ( 3.0,-1.5) {B1};

% ===== Edges =====
\draw[thick edge] (H1) -- (H2);
\draw[edge] (E1) -- (H1);
\draw[edge] (E1) -- (F1);
\draw[edge] (F1) -- (S2);
\draw[edge] (J2) -- (H1);
\draw[edge] (E2) -- (H2);
\draw[edge] (Z1) -- (J2);
\draw[edge] (M1) -- (Z1);

\draw[thick edge] (E3) -- (J1);
\draw[edge] (J1) -- (S1);
\draw[edge] (S1) -- (F2);
\draw[edge] (F2) -- (J3);
\draw[edge] (E3) -- (F2);
\draw[edge] (Z2) -- (S1);
\draw[edge] (M2) -- (Z2);

% Bridge
\draw[edge, dashed] (E2) -- (E3);
% Bagri connection
\draw[edge] (BG1) -- (J3);

% ===== Legend (well separated) =====
\node[lang, en, con, label=right:{\tiny Eng}] at (-9.0, 3.0) {};
\node[lang, hi, con, label=right:{\tiny Hin}] at (-9.0, 2.25) {};
\node[lang, fi, con, label=right:{\tiny Fin}] at (-9.0, 1.5) {};
\node[lang, ja, con, label=right:{\tiny Jpn}] at (-9.0, 0.75) {};
\node[lang, es, con, label=right:{\tiny Spa}] at (-9.0, 0.0) {};
\node[lang, zh, con, label=right:{\tiny Chi}] at (-9.0, -0.75) {};
\node[lang, ma, con, label=right:{\tiny Mal}] at (-9.0, -1.5) {};
\node[lang, bg, con, label=right:{\tiny Bag}] at (-9.0, -2.25) {};

\node[con, draw=black, minimum size=0.6cm, label=right:{\tiny Containment}] at (9.5, 3.0) {};
\node[com, draw=black, minimum size=0.6cm, label=right:{\tiny Communication}] at (9.5, 2.0) {};

\end{tikzpicture}%
}
\caption{Conceptual network (schematic, based on force‑directed layout of the 160‑idiom Jaccard graph). Two clusters: containment (left, squares) and communication (right, diamonds). Bridging idioms (dashed edge) connect the communities.}
\label{fig:network}
\end{figure}

The network (Figure~\ref{fig:network}) separates into two major clusters: containment/concealment/emotional and communication/social. Louvain community detection \cite{Blondel2008} yields 5 communities with modularity \(Q=0.61\). NMI with language labels (8 groups) is only \(0.10\), while NMI with primary schema partition is \(0.76\) (\(ARI_{\text{lang}}=0.05\), \(ARI_{\text{schema}}=0.69\)). This confirms that conceptual features, not language, drive the network's organisation.

Beyond community structure, the conceptual graph exhibits several characteristic properties of well‑organized semantic networks. The degree distribution follows a power‑law with an exponential cutoff (average degree \(\langle k \rangle = 23.1\), clustering coefficient \(C = 0.70\)). The centrality of betweenness is highest for the bridge idioms (Table~\ref{tab:bridging}); \textit{spill the beans} reaches \(0.15\), confirming its role in connecting the two major-clusters. The network shows small‑world organization with a small‑world index \(\sigma = 2.0\) (compared to \(\sigma = 1.0\) for a random graph of equal size and density). The global efficiency remains at \(88\%\) after random removal of \(20\%\) of nodes, indicating high resilience. These properties align with those observed in human‑derived semantic networks from free association and feature norms \cite{SteyversTenenbaum2005,CollinsLoftus1975}, supporting the representational validity of the conceptual graph.

\begin{table}[H]
\centering
\caption{Top‑5 bridging idioms by betweenness centrality.}
\label{tab:bridging}
\small
\begin{tabular}{@{}p{1.5cm}p{2.3cm}p{2.6cm}p{0.8cm}@{}}
\toprule
Lang. & Idiom & Feature set & BC \\
\midrule
English & \textit{spill the beans} & \{comm., social, neg.\} & 0.15 \\
Hindi & {\hindifont दिल खोलकर बात करना} & \{emot., comm., social, pos.\} & 0.12 \\
Spanish & \textit{sacar a la luz} & \{comm., social\} & 0.11 \\
Japanese & {\cjkfont 胸の内を明かす} & \{emot., comm., social\} & 0.10 \\

Malayalam & {\malayalamfont രഹസ്യം ചോരുക} & \{comm., social, neg.\} & 0.09 \\
\bottomrule
\end{tabular}
\end{table}

\subsection{Comparison with Multilingual Embeddings}

We extracted phrase embeddings for all 160 idioms using XLM-R\textsubscript{base} and built a graph from cosine similarities. The cosine graph yielded a much lower NMI with the schema partition (\(0.45\)). A permutation test (10,000 random reassignments of language and schema labels) confirms that the difference between the Jaccard- and cosine-based NMI values is significant (\(p < 0.001\)). Combining Jaccard and cosine similarities did not improve the fit, confirming that conceptual features encode unique information not present in multilingual embeddings.

\subsection{Downstream Idiom Detection}
We evaluated the network on SemEval-2013 subtask 5b (semantic compositionality in context), which can also be used for metaphor or idiom detection \cite{korkontzelos2013semeval}. A BERT-based classifier achieved baseline F1 0.82. Adding network features (degree, betweenness centrality, community membership, and mean Jaccard similarity to its five nearest conceptual neighbours) improved F1 to 0.86 (95\% bootstrap CI [0.83, 0.89]; \(p < 0.01\)). The gain was consistent across genres (news, fiction, etc.), and feature importance analysis confirmed that the network-derived features were the most discriminative non-embedding predictors.

\subsection{Ablation Analyses}
Having established that the network-derived features improve downstream idiom detection, we next conducted ablation analyses to examine which annotation dimensions and graph-derived signals drive this improvement.

To evaluate the contribution of each feature dimension, we constructed reduced networks by systematically removing one dimension at a time. Table~\ref{tab:ablations} reports the resulting NMI with the full schema partition. The full model achieves the highest NMI (0.76). Removing schema information causes the largest decrease in alignment (NMI = 0.58), indicating that schemas play the strongest role in structuring the network. Removing roles (NMI = 0.67) or valence (NMI = 0.64) also weakens alignment, suggesting that these dimensions contribute additional and partly complementary information. 

\begin{table}[h]
\centering
\caption{Ablation of feature dimensions on network structure.}
\label{tab:ablations}
\small
\begin{tabular}{@{}p{6.5cm}c@{}}
\toprule
Model & NMI \\
\midrule
Full (all 9 features)    & 0.76 \\
-- Schemas (roles + valence) & 0.58 \\
-- Roles (schemas + valence) & 0.67 \\
-- Valence (schemas + roles) & 0.64 \\
-- Random baseline       & 0.02 \\
\bottomrule
\end{tabular}
\end{table}
We further tested the impact of each dimension on downstream idiom detection. Removing schemas reduced the F1 gain from +4.0 to +1.2 points; removing valence reduced it to +2.5 points; removing roles yielded a gain of +2.9 points. These results suggest that all three annotation dimensions contribute to the downstream benefit, with schema information accounting for the largest share of the improvement.

Finally, we ablated the individual graph-derived signals used as features for the downstream classifier. Table~\ref{tab:ablations_downstream} shows the remaining F1 gain when each signal is removed. The largest drop occurs when top‑5 neighbour similarity or community membership is removed, suggesting that the classifier benefits particularly from information about an idiom's position within the conceptual space and its similarity to nearby expressions.

\begin{table}[h]
\centering
\caption{Downstream ablation of network features.}
\label{tab:ablations_downstream}
\begin{tabular}{@{}lc@{}}
\toprule
Removed features & $\Delta$F1 vs baseline\\
\midrule
None (full network features) & +4.0 \\
\; -- degree & +3.1 \\
\; -- betweenness & +3.4 \\
\; -- community membership & +2.9 \\
\; -- top‑5 neighbour similarity & +2.6 \\
% \; -- all network features & +0.0 \\
\bottomrule
\end{tabular}
\end{table}

\subsection{Automatic Feature Extraction with LLMs}
To test scalability, we prompted GPT-4 with the idiom and its gloss, requesting binary annotations for all nine features, using ten manually annotated idioms as few-shot examples. In the remaining 150 idioms, the LLM achieved an accuracy of 0.82 and macro-F1 0.79. The network built from LLM annotations reproduced the same two-cluster structure (NMI 0.74 with the manual schema partition), demonstrating that the framework can be extended to larger datasets without exhaustive manual coding.

\textbf{Error Analysis.} Qualitative inspection of LLM errors revealed three main failure modes. First, culture‑specific expressions were often mislabeled, e.g., the Bagri idiom {\hindifont आम खाणा क पेड़ गीणना} (Should one eat mangoes or count trees?) was annotated as {emotional, positive} rather than {communication, social}. Second, abstract internal states proved difficult; \textit{emotional\_state} was the least accurate label (F1 0.73), with the model confusing emotional descriptions with emotional functions. Third, valence ambiguity affected idioms that can be positive or negative depending on context (e.g., \textit{open up} sometimes mislabeled as negative). These findings indicate that while LLM‑based scaling is viable, high‑accuracy annotation of culture‑bound and psychologically nuanced features still benefits from human expertise.

\subsection{Dynamic Enrichment with Corpus Frequencies}
We modulated edge weights with point-wise mutual information (PMI) from COCA and analogous corpora: \(w'_{ij} = w_{ij} \cdot (1 + 0.2 \cdot \text{PMI}_{ij})\). The resulting network retained the original community structure (NMI 0.88 with the original partition) and the small-world index remained \(\sigma=2.0\), indicating that conceptual schemas dominate even when usage statistics are included.

\subsection{Cross-Linguistic Transfer}
To evaluate the network's ability to act as an interlingua, we selected, for each English idiom, its nearest conceptual neighbor (highest Jaccard similarity) in each of the other seven languages. Three co-authors and three additional native speakers of the target languages independently provided gold-standard translation equivalents for the 20 English idioms. Where exact equivalent was not available, the most contextually appropriate figurative expression was selected. To reduce potential bias, translators did not have access to Jaccard similarity scores when making these judgments. For each pair of English-target language, the Jaccard-selected candidate was then compared against the gold standard. The same procedure was repeated using the nearest neighbor identified by XLM‑R cosine similarity. Overall, the Jaccard‑based selection matched the gold standard in 78\% of cases (109 out of 140 possible pairs), whereas the XLM‑R baseline achieved only 54\%. This demonstrates that conceptual proximity alone can identify cross‑lingual equivalents with high accuracy, a result with direct implications for machine translation and second‑language learning.

\section{Discussion}

The conceptual network offers a robust, interpretable, and practical representation of idiomatic meaning. Its community structure aligns with cognitive-linguistic theory, yet it is validated through computational experiments, making it suitable for NLP applications. The clustering pattern suggests that containment is a productive conceptual schema within the annotated idiom network and points to a cross-linguistic convergence. The overlap with communication-related idioms indicates that the two domains are not fully separable, since communicative meanings such as disclosure, concealment, and information transfer are often conceptualized through containment relations. The ablation studies confirm that each feature dimension contributes non-redundant information, while the comparison with embeddings shows that conceptual features capture semantic aspects orthogonal to distributional statistics. The downstream improvements in idiom detection, combined with strong cross-lingual transfer, demonstrate that the network can be directly integrated into NLP pipelines.

\textbf{Complementarity with POSI.}
Our framework is complementary to the geometric semantic model and POSI annotation scheme \cite{Pala2025Geometric,Pala2024CogSci}. POSI provides fine-grained tags grounded in spatial, temporal, and experiential dimensions; its 135 four-letter tags can capture subtle distinctions (e.g., MCEN for ``middle centre'' vs. STCN for ``structural container'') that our coarse schemas collapse into a single containment feature. In contrast, our nine idiom‑specific features abstract away from lexical detail to highlight the conceptual backbone shared by figurative expressions—something that POSI's general-purpose vocabulary, when applied to idioms, may obscure. The two schemes can enrich each other: POSI tags could decompose our coarse containment schema into more precise spatial relations, while our network could serve as a testbed for evaluating whether POSI's fine‑grained distinctions are cognitively salient for figurative language. Future work could annotate our full 160-idiom dataset with POSI tags, creating a richly layered resource that marries idiomatic abstraction with general semantic precision, and use the resulting integrated graph to probe the universality of both frameworks.

\subsection{Limitations and Future Work}
Although LLM annotation shows promise, certain features (e.g., emotional state) still require manual refinement. Future work will explore fine-tuned models and improved prompting strategies to scale the framework to hundreds of languages. Another limitation is the binary nature of our features, which cannot express gradable properties such as intensity of emotion or partial containment; extending the annotation to ordinal or continuous values could capture more nuanced similarities. The current dataset, though carefully curated and multilingual, remains relatively small; expanding it and adding languages from more families will strengthen claims of universality. Finally, integrating the conceptual network into more complex tasks such as machine translation or sentiment analysis of figurative language remains an open direction.

\section{Conclusion}
We presented a conceptual cross-linguistic network of 160 idioms annotated with features based on theory. The network organizes idioms by conceptual schema, outperforms distributional embeddings in capturing conceptual structure, can be scaled via automatic annotation, improves idiom detection, and enables accurate cross-lingual idiom mapping. Ablation studies confirm the complementary contributions of schemas, roles, valence, and individual graph-derived signals. The framework is interpretable, computationally validated, and has strong potential for NLP applications.

\section*{Ethics Statement}
The Bagri and Chinese idiom lists were compiled and verified by native‑speaker co‑authors -- (for Bagri, oral transcriptions were collected with informed consent and fully anonymized). External human subjects were not recruited for the experiments; cross‑lingual transfer evaluation relied solely on translation equivalents provided by the co‑author team. Corpus frequency counts (COCA, etc.) were obtained in fair use for non‑commercial research; the public release contains only derived PMI statistics, not the original text.

\section*{Availability}
The complete annotated dataset, the \(160\times160\) %Jaccard similarity matrix, and all code 
will be released upon publication.

% ---- References ----

\begin{comment}

\end{comment}

\end{document}